\documentclass{article}

\usepackage{microtype}
\usepackage{graphicx}
\usepackage{subfigure}
\usepackage{booktabs} 
\usepackage{sidecap}
\usepackage{subcaption}
\usepackage{multirow}
\usepackage{multicol}
\usepackage{stfloats}

\usepackage{hyperref}

\usepackage[accepted]{icml2025}

\usepackage{amsmath}
\usepackage{amssymb}
\usepackage{mathtools}
\usepackage{amsthm}
\usepackage[capitalize,noabbrev]{cleveref}

\theoremstyle{plain}
\newtheorem{theorem}{Theorem}[section]

\theoremstyle{definition}
\newtheorem{definition}[theorem]{Definition}

\theoremstyle{remark}

\usepackage[textsize=tiny]{todonotes}
\usepackage{hyperref}
\usepackage{url}
\usepackage{float}
\usepackage{amssymb}
\usepackage{amsmath}
\usepackage{amsthm}
\usepackage{enumitem}
\usepackage{caption}
\usepackage{multirow}
\usepackage{graphicx}
\usepackage{xcolor}

\usepackage{makecell}

\DeclareUnicodeCharacter{2212}{\ensuremath{-}}

\usepackage{algorithm,algorithmic}

\def\sup{\displaystyle\mathop {\mbox{\rm sup}}}

\newcommand{\bu}{{\bf u}}

\newcommand{\bbp}{{\bf P}}

\newcommand{\bba}{{\bf A}}
\newcommand{\bbb}{{\bf B}}
\newcommand{\bbq}{{{\bf Q}}}
\newcommand{\bbr}{{{\bf R}}}
\usepackage{comment}

\newcommand{\bbs}{{\bf S}}

\newcommand{\bbu}{{\bf U}}

\newcommand{\bbg}{{\bf G}}

\newcommand{\bbv}{{\bf V}}
\newcommand{\bbx}{{\bf X}}
\newcommand{\bby}{{\bf Y}}

\newcommand{\bbw}{{\bf W}}

\newcommand{\bbm}{{\bf M}}

\makeatletter
\newcommand*{\rom}[1]{\expandafter\@slowromancap\romannumeral #1@}
\makeatother

\newcommand{\diag}{\mbox{diag}}

\usepackage{wrapfig}
\usepackage{multicol}

\usepackage{tcolorbox} 
\tcbuselibrary{listingsutf8}

\icmltitlerunning{LORENZA: Enhancing Generalization in Low-Rank Gradient LLM Training via Efficient Zeroth-Order Adaptive SAM}
\usepackage{amsthm}
\begin{document}

\twocolumn[
\icmltitle{LORENZA: Enhancing Generalization in \underline{Lo}w-\underline{R}ank Gradient LLM Training and Fine-Tuning via \underline{E}fficie\underline{n}t \underline{Z}eroth-Order {\underline{A}}daptive SAM Optimization}

\begin{icmlauthorlist}
\icmlauthor{Yehonathan Refael}{tau}
\icmlauthor{Iftach Arbel}{tau}
\icmlauthor{Ofir Lindenbaum}{biu}
\icmlauthor{Tom Tirer}{biu}
\end{icmlauthorlist}

\icmlaffiliation{biu}{Faculty of Engineering, Bar-Ilan University, Ramat-Gan 5290002, Israel.}
\icmlaffiliation{tau}{Faculty of Engineering, Tel Aviv University, Tel Aviv 6997801, Israel}

\icmlcorrespondingauthor{Yehonathan Refael}{refaelkalim@mail.tau.ac.il}
\icmlkeywords{Machine Learning, ICML}

\vskip 0.3in
]

\printAffiliationsAndNotice{}
\begin{abstract}
We study robust parameter-efficient fine-tuning (PEFT) techniques designed to improve accuracy and generalization while operating within strict computational and memory hardware constraints, specifically focusing on large-language models (LLMs). Existing PEFT methods often lack robustness and fail to generalize effectively across diverse tasks, leading to suboptimal performance in real-world scenarios. To address this, we present a new highly computationally efficient framework called AdaZo-SAM, combining Adam and Sharpness-Aware Minimization (SAM) while requiring only a single-gradient computation in every iteration. This is achieved using a stochastic zeroth-order estimation to find SAM's ascent perturbation. We provide a convergence guarantee for AdaZo-SAM and show that it improves the generalization ability of state-of-the-art PEFT methods. Additionally, we design a low-rank gradient optimization method named LORENZA, which is a memory-efficient version of AdaZo-SAM. LORENZA utilizes a randomized SVD scheme to efficiently compute the subspace projection matrix and apply optimization steps onto the selected subspace. This technique enables full-parameter fine-tuning with adaptive low-rank gradient updates, achieving the same reduced memory consumption as gradient-low-rank-projection methods. We provide a convergence analysis of LORENZA and demonstrate its merits for pre-training and fine-tuning LLMs.

\end{abstract}

\section{Introduction}
Large language models (LLMs) have attracted considerable attention due to their remarkable ability to perform various tasks, such as engaging in dialogue and completing text. Their performance can be improved through supervised fine-tuning and additional pre-training across different tasks and domains. However, training these models presents substantial computational power and memory challenges. This difficulty arises because the process of updating gradients necessitates storing billions of trainable parameters along with the optimizer's state (which includes gradients and moments). For instance, in the Adam optimizer \citep{kingma2017adammethodstochasticoptimization}, the storage requirements for gradients and the estimated first and second moments can triple the overall size of the model \citep{xu2024understanding,brown2022efficient,kim2023memory}. 

Researchers have developed various optimization techniques to reduce memory usage during model training to tackle the challenges associated with LLM fine-tuning. One key research topic that has emerged is Parameter-Efficient Fine-Tuning (PEFT) \citep{han2024parameterefficientfinetuninglargemodels}, which enables the adaptation of pre-trained language models to different tasks without the need to fine-tune all model parameters. A prominent method within PEFT is the Low-Rank Adaptation (LoRA) algorithm, introduced by \cite{hu2021lora}. LoRA reparameterizes a weight matrix $ \bbw \in \mathbb{R}^{m \times n} $ into $ \bbw = \bbw_0 + \bbb \bba $, where $ \bbw_0 $ is a frozen full-rank matrix, and $ \bbb \in \mathbb{R}^{m \times r} $ and $ \bba \in \mathbb{R}^{r \times n} $ are low-rank adaptors. Since $ r \ll \min(m, n) $, the low-rank adaptors $ \bba $ and $ \bbb $ require fewer trainable parameters, reducing memory usage. LoRA has been widely adopted for fine-tuning, with many variants emerging, such as \cite{Chen2023,Xu2023,Wang2023}.

\begin{table*}[hbt]
\caption{Comparison between LORENZA, GaLore \cite{zhao2024galore}, LoRA \cite{hu2021lora}, SAM \cite{foret2021sam}, and AdaSAM \cite{sun2023adasamboostingsharpnessawareminimization}. Assume $\bbw \in \mathbb{R}^{n \times m} (n \geq m)$, constant low-rank $r$.}
\centering
\footnotesize
\begin{tabular}{lccccc}
\hline
& LORENZA & GaLore & LoRA & SAM & AdaSAM\\
\hline
Weights & $n m$ & $n m$ & $n m + {\color{red}n r + m r}$ & $n m$ & $n m$\\
\hline
Optim States & $n r + 2 m r$ & $n r + 2 m r$ & ${\color{red}2} n r + 2 m r$ & ${\color{red}n m}$ & ${\color{red}4n m}$\\
\hline
Computation  & $O\left(m n r + mnr / T\right)$ & $O\left(m n r+m^{\color{red}2} n / T\right)$ & $O\left(m nr\right)$ & $O\left(m nr\right)$ & $O\left(m nr\right)$ \\
\hline
Fine-Tuning & ${\color{green}\checkmark}$ & ${\color{green}\checkmark}$ & ${\color{green}\checkmark}$ & ${\color{green}\checkmark}$ & ${\color{green}\checkmark}$ \\
Pre-Training & ${\color{green}\checkmark}$ & ${\color{green}\checkmark}$ & ${\color{red}\mathbf{x}}$ & ${\color{green}\checkmark}$ & ${\color{green}\checkmark}$ \\
Multi-Subspace & ${\color{green}\checkmark}$ & ${\color{green}\checkmark}$ & ${\color{red}\mathbf{x}}$ & ${\color{red}\mathbf{x}}$ & ${\color{red}\mathbf{x}}$\\
Num. of Backprop per step & ${\color{green}1}$ & ${\color{green}1}$ & ${\color{green}1}$ & ${\color{red}2}$ & ${\color{red}2}$\\
Sharpness-Aware & ${\color{green}\checkmark}$ & ${\color{red}\mathbf{x}}$ & ${\color{red}\mathbf{x}}$ & ${\color{green}\checkmark}$ & ${\color{green}\checkmark}$\\
\hline
\end{tabular}
\end{table*}
Despite its advantages, recent research has identified some limitations of low-rank reparameterization. One notable weakness of LoRA-type methods is their potential to fall short in recovering accuracy for more challenging fine-tuning tasks compared to Full Fine-Tuning (FFT) \citep{meng2024periodiclorabreakinglowrankbottleneck}. These issues may stem from the fact that optimal weight matrices are not inherently low-rank or from changes in gradient training dynamics introduced by the reparameterization. In an effort to mitigate this gap, recently, a low-rank adaptation method, GaLore \cite{zhao2024galore}, has been proposed to fine-tune LLMs efficiently by updating the model weights within a low-rank subspace. This method significantly reduces the number of fine-tuned parameters, offering several key advantages. First, it eliminates the need for additional adapters alongside the pre-trained model. Second, it removes the requirement to store all gradient parameters during training. Third, it reduces memory usage by bypassing the need to retain optimizer states. Demonstrating its effectiveness, GaLore successfully pre-trained an LLM with 7 billion parameters on a consumer-level GPU with just 24GB of memory. Naturally, the method garnered significant attention, leading the study of several variations aimed at further reducing memory consumption \citep{refael2025adarankgrad,liao2024,zhang2024qgalore,das2024,huang2024galoremini}.

A notable limitation of low-rank type methods is their tendency to struggle in reaching the performance of FFT in challenging tasks (see Table \ref{tab:comparison}) and to generalize well in out-of-sample, domain shift, and zero-shot tasks. This performance disparity tends to emerge in scenarios involving complex target tasks, such as mathematical reasoning or coding. This raises an open question: Can a PEFT method be developed that offers the practical advantages and simplicity of LoRA-like techniques while matching the superior accuracy of FFT?

To address this question, we introduce LORENZA, a computationally efficient, sharpness-aware optimization method that leverages adaptive low-rank gradient updates and memory-efficient zeroth-order sharpness minimization to enhance generalization. Unlike existing sharpness-aware fine-tuning methods that require costly double backpropagation, LORENZA eliminates this overhead through a backpropagation-free perturbation (BPFP) scheme, significantly reducing computational and memory complexity. For example, for the OPT-13B model, the backpropagation consumes approximately $\times 6$ more memory during fine-tuning than using the proposed alternative BPFP and approximately $\times 4$ more in calculation time. Furthermore, LORENZA employs a dynamic low-rank subspace selection mechanism, ensuring optimization updates remain efficient while maintaining the benefits of full-rank tuning.

From a theoretical perspective, we establish a convergence guarantee for LORENZA, proving that it efficiently finds flat minima that promote better generalization while maintaining computational efficiency. Our analysis shows that LORENZA retains the key benefits of adaptive sharpness-aware minimization while significantly reducing memory overhead.

Empirically, we demonstrate that LORENZA outperforms state-of-the-art methods in both pre-training and fine-tuning of LLMs. Specifically, LORENZA achieves higher accuracy and improved generalization on a variety of fine-tuning benchmarks, outperforming existing low-rank adaptation techniques such as LoRA and GaLore \cite{hu2021lora,zhao2024galore} while significantly reducing memory consumption. Additionally, LORENZA exhibits robust adaptation across diverse datasets and challenging tasks, demonstrating its potential as a scalable, efficient alternative for training and fine-tuning LLMs under resource constraints.
 
\section{Related work}

\begin{figure}[htb]
\centering
\includegraphics[width=0.95\linewidth]{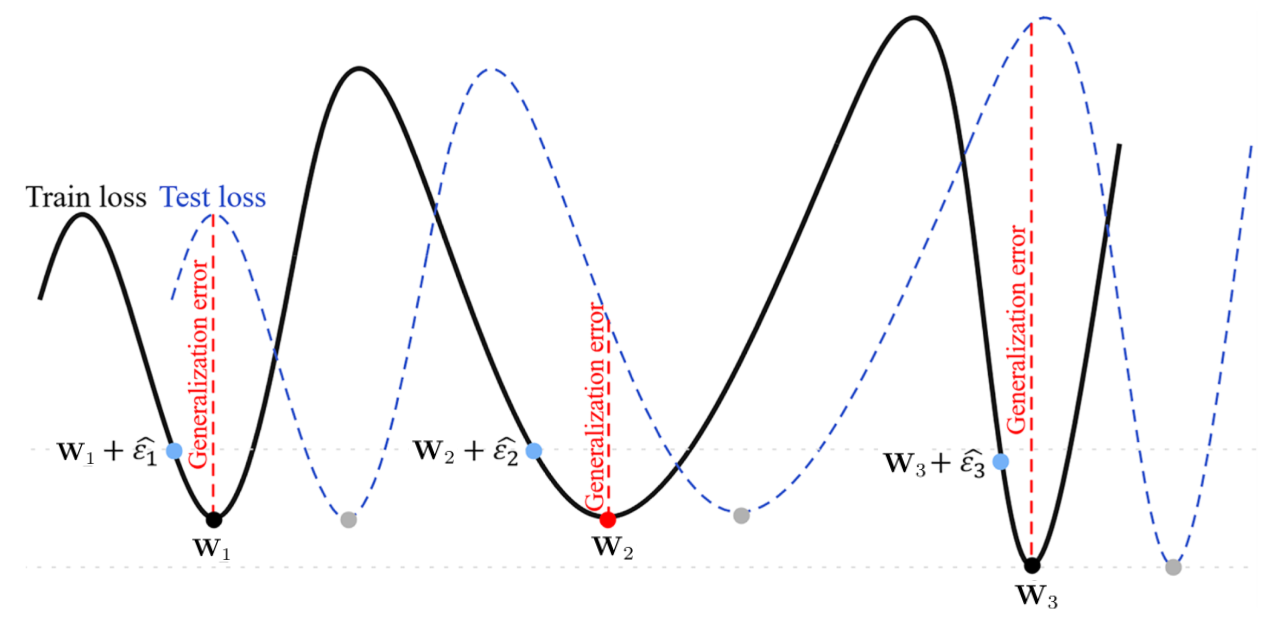} 
\caption{An illustration showing how the flatness of different minima can impact test loss. Specifically, $\bbw_1$ and $\bbw_3$, are located in sharp regions that have a high generalization error, while $\bbw_2$, found in a flatter region, exhibits a lower generalization error \citep{rs16162877}.} 
\label{fig:sam}
\end{figure}

The generalization ability of neural networks has been shown to correlate with the flatness of the minima \citep{hochreiter1997flat,keskar2017largebatchtrainingdeeplearning,sun2023adasamboostingsharpnessawareminimization,si2023,yue2024sharpnessawareminimizationrevisitedweighted}. 
In regions around flat minima in the loss landscape, as illustrated in Figure \ref{fig:sam}, small parameter changes lead to minimal loss variation, reducing the model's sensitivity to noise and perturbations. This robustness has been shown to enhance the model's ability to generalize to unseen data, compared to standard optimization methods that may converge to sharp minima.

\paragraph{Sharpness-Aware Minimization.}  
Sharpness-Aware Minimization (SAM)~\cite{foret2021sam} aims to solve the min-max optimization problem: $\min_{\mathbf{w}} \max_{\|\boldsymbol{\epsilon}\|_2 \leq \rho} f_S(\mathbf{w} + \boldsymbol{\epsilon})$, where $f_S(\mathbf{w})$ denotes the empirical loss.
Note that the objective value per $\mathbf{w}$ is equal to the highest value of the loss within a neighborhood of $\mathbf{w}$, defined as a ball of radius $\rho$ centered at $\mathbf{w}$.
Therefore, this problem promotes flat minimizers, where small perturbations in the weights (even the ``worst'' $\boldsymbol{\epsilon}$) do not increase the empirical loss significantly. 

To simplify the problem,
SAM approximates the solution to the inner maximization using a first-order Taylor expansion around $\mathbf{w}$. This leads to the following approximation of the perturbation $\boldsymbol{\epsilon}$, 
\[{\footnotesize\boldsymbol{\epsilon} = \arg\max_{\|\boldsymbol{\epsilon}\|_2 \leq \rho} f_S(\mathbf{w} + \boldsymbol{\epsilon}) \approx \rho \frac{\nabla_{\mathbf{w}} f_S(\mathbf{w})}{\|\nabla_{\mathbf{w}} f_S(\mathbf{w})\|_2}}.\]  
Substituting this back into the outer minimization reformulates the objective as:  
${\footnotesize
\min_{\mathbf{w}} f_S\left(\mathbf{w} + \rho \frac{\nabla_{\mathbf{w}} f_S(\mathbf{w})}{\|\nabla_{\mathbf{w}} f_S(\mathbf{w})\|_2}\right)}.$  
In practice, given a mini-batch $B$, SAM extends the standard stochastic gradient descent (SGD) \cite{battash2024revisiting} update to the following two-step process: $(1)$ for the current weights $\mathbf{w}_t$, compute the adversarial perturbation:  
$\boldsymbol{\epsilon}_t = \rho \frac{\nabla_{\mathbf{w}} f_B(\mathbf{w}_t)}{\|\nabla_{\mathbf{w}} f_B(\mathbf{w}_t)\|_2},$  
(2) evaluate the gradient of the perturbed weights $\mathbf{w}_t + \boldsymbol{\epsilon}_t$ and use it to update $\mathbf{w}_t,$ namely  
$\mathbf{w}_{t+1} = \mathbf{w}_t - \eta \bbg^{\text{SAM}}_t,$ where $\bbg^{\text{SAM}}_t = \nabla_{\mathbf{w}} f_B(\mathbf{w}_t + \boldsymbol{\epsilon}_t),$ and $\eta$ is the learning rate. 
This procedure ensures that SAM balances the trade-off between minimizing the empirical loss and achieving a flat minimum, improving generalization performance.

\paragraph{AdaSAM.}
AdaSAM~\cite{sun2023adasamboostingsharpnessawareminimization} enhances SAM by integrating adaptive estimates of the first and second moments of the gradients to further improve optimization efficiency and generalization in deep neural networks, similar to the popular Adam optimizer~\citep{kingma2017adammethodstochasticoptimization} and its weight decay regularization variant, AdamW~\citep{loshchilov2019decoupledweightdecayregularization}. Specifically, the algorithm alternates between calculating a perturbation and updating parameters using the Adam optimization rule. Formally, AdaSAM modifies the SAM optimization process by incorporating the notion of AdaM~\cite{tan2019convergence}, introducing a momentum term,
$\bbm_t = \beta_1 \bbm_{t-1} + (1 - \beta_1)\bbg^{\text{SAM}}_t,$
which is weighted by a momentum factor \(\beta_1\). Additionally, it tracks a second-moment estimate using a smoothing parameter \(\beta_2\), namely
$\bbv_t = \beta_2 \bbv_{t-1} + (1 - \beta_2)\left[\bbg^{\text{SAM}}_t\right]^2.$ This allows it to dynamically adjust using historical gradient information.

AdaSAM achieves a convergence rate of $\mathcal{O}(1 / \sqrt{bT})$, where $b$ is the batch size, providing a linear speedup with increased batch sizes, making it suitable for large-scale training scenarios.

This adaptive variant of SAM requires the storage of both $\bbm_t$ and $\bbv_t$ at each time step, resulting in a memory cost of $2mn$. Additionally, the perturbation introduces an extra memory usage of $mn$, bringing the total memory access for AdaSAM optimization to $4mn$. It is important to note that this inefficiency not only leads to high memory requirements but also increases computational time. Compared to gradient-based optimizers, SAM and its variants involve two gradients, which means two backpropagation procedures are performed during a single update step.

\paragraph{Surrogate Gap Guided Sharpness-Aware Minimization (GSAM).}  
GSAM \cite{zhuang2022surrogate} extends SAM by jointly minimizing the perturbed loss \( f_p(\mathbf{w}) \) and the surrogate gap \( h(\mathbf{w}) = f_p(\mathbf{w}) - f(\mathbf{w}) \), which measures sharpness. The algorithm first computes the perturbed weight \( \mathbf{w}_{\text{adv}} = \mathbf{w} + \rho \frac{\nabla f(\mathbf{w})}{\|\nabla f(\mathbf{w})\|_2} \), where \( \rho \) defines the neighborhood radius, and evaluates the gradient \( \nabla f_p(\mathbf{w}) = \nabla f(\mathbf{w}_{\text{adv}}) \). To reduce the surrogate gap without affecting the minimization of the perturbed loss, the gradient \( \nabla f(\mathbf{w}) \) is decomposed into parallel and orthogonal components with respect to \( \nabla f_p(\mathbf{w}) \), expressed as \( \nabla f(\mathbf{w}) = \nabla_\parallel f(\mathbf{w}) + \nabla_\perp f(\mathbf{w}) \), and only $\nabla_\perp f(\mathbf{w})$ is utilized for minimizing $h(\mathbf{w})$. Specifically, the final update adjusts the weights to \( \mathbf{w}_{t+1} = \mathbf{w}_t - \eta \left( \nabla f_p(\mathbf{w}_t) - \alpha \nabla_\perp f(\mathbf{w}_t) \right) \), where \( \alpha \) controls the term that promotes minimization of $h(\mathbf{w})$. This approach ensures the model converges to a flat minimum with better generalization by maintaining low \( f_p(\mathbf{w}) \) while explicitly reducing a measure of sharpness.

\paragraph{Memory efficient optimizers.}
Recently, several works have focused on developing memory-efficient optimization techniques. Multiple studies have aimed to reduce the memory requirements of gradient statistics in adaptive optimization algorithms \citep{shazeerAdafactorAdaptiveLearning,anilMemoryEfficientAdaptive}. One common approach is quantization, which helps decrease the memory footprint of optimizer states \citep{liMemoryEfficientOptimizers2023}. Additionally, recent advancements have suggested reducing the memory used by weight gradients by integrating the backward operation with the optimizer update \citep{lvAdaLomoLowmemoryOptimization2023,lvFullParameterFinetuning2023}. This characteristic has been leveraged to reduce memory usage during training processes \citep{gooneratneLowrankGradientApproximation2020, huangLowRankGradientDescent2023, modoranuErrorFeedbackCan2024}.
Efforts to reduce SAM's memory demands have been reported as well. They all appear to focus solely on the overhead caused by the perturbation (ascent step) computation. FSAM \cite{Zhong2022ImprovingSM} and SSAM \cite{zhao2023randomizedsharpnessawaretrainingboosting} leverage Fisher information to selectively perturb a parameter subset, achieving 50\%-90\% memory savings from the overhead at the cost of increased computation. Recent work on $\nu$-SAM \cite{anonymous2024sam} employs nuclear norm constraints during the ascent step for greater memory efficiency. Similarly, SAM-ON \cite{mueller2023} focuses perturbations solely on normalization layers. However, these approaches do not address the memory complexity of the baseline optimizer performing the descent step.
Furthermore, they often trade memory savings related to the ascent step with increased computational complexity 
or struggle to generalize across diverse fine-tuning tasks. 
Our method bridges this gap by introducing a low-rank gradient optimization framework that is applied in both ascent and descent directions. We also estimate randomized ascent direction (gradient perturbation), leading to both memory efficiency and computational simplicity while enabling robust generalization in pre-training and fine-tuning of LLMs.

\paragraph{Low-rank gradient optimization.} The phenomenon of low-rank gradients naturally arises during the training of neural networks, a subject that has been extensively examined both theoretically and practically, e.g., \cite{zhaoZerOInitializationInitializing2022, cossonLowRankGradientDescent2023, yang2023spectral}. This characteristic low-rank structure gradient has been leveraged to reduce memory usage during training processes \cite{gooneratneLowrankGradientApproximation2020, huangLowRankGradientDescent2023, modoranuErrorFeedbackCan2024}, and results in a reduced computational complexity as compared to standard gradient descent methods. Recent work in \cite{refael2025adarankgrad} theoretically and empirically showed a natural phenomenon in which the rank of reversible layer gradients \citep{tian2021} monotonically diminishes to one during training and suggested to leverage to adaptively reduces the rank of the gradients during Adam optimization steps.

Recent works by \cite{zhao2024galore,refael2025adarankgrad} suggest optimization methods reducing the cost of Adam's states (first and second-order moments) by projecting the model gradients on a most meaningful low subspace, thus inherently, the optimizer's state gets a very low dimensionality structure. Both works introduced a mechanism for updating the subspace onto which gradients are projected, enabling them to establish convergence guarantees. While these methods effectively reduce memory consumption and perform well on relatively homogeneous tasks, they struggle to maintain accuracy on more complex challenges, such as reasoning or coding, when compared to FFT. Additionally, they do not generalize effectively to out-of-sample scenarios, domain shifts, or zero-shot tasks.

\paragraph{Zeroth-Order Optimization} Zeroth-order (ZO) optimization estimates gradients using finite differences and relies only on function value oracles. Despite this, its structure is similar to first-order (FO) gradient-based methods. It has gained significant attention due to its effectiveness across various modern machine learning challenges \citep{liu2020primer}.

Methods for ZO include approaches that leverage historical data to enhance gradient estimators \citep{meier2019improving, cheng2021convergence}. These methods utilize gradient structural information \citep{singhal2023guess}, exploit sparsity to reduce dimensional dependence \citep{cai2021zeroth, cai2022zeroth, chen2023deepzero}, and reuse intermediate features \citep{chen2023deepzero} or random perturbations \citep{malladi2024finetuning}. These strategies have shown significant advancements in addressing large-scale machine-learning challenges. In this study, we will further leverage the effectiveness of ZO to reduce unnecessary computational costs.

\begin{figure}[htb]
\centering
\includegraphics[width=0.8\linewidth]{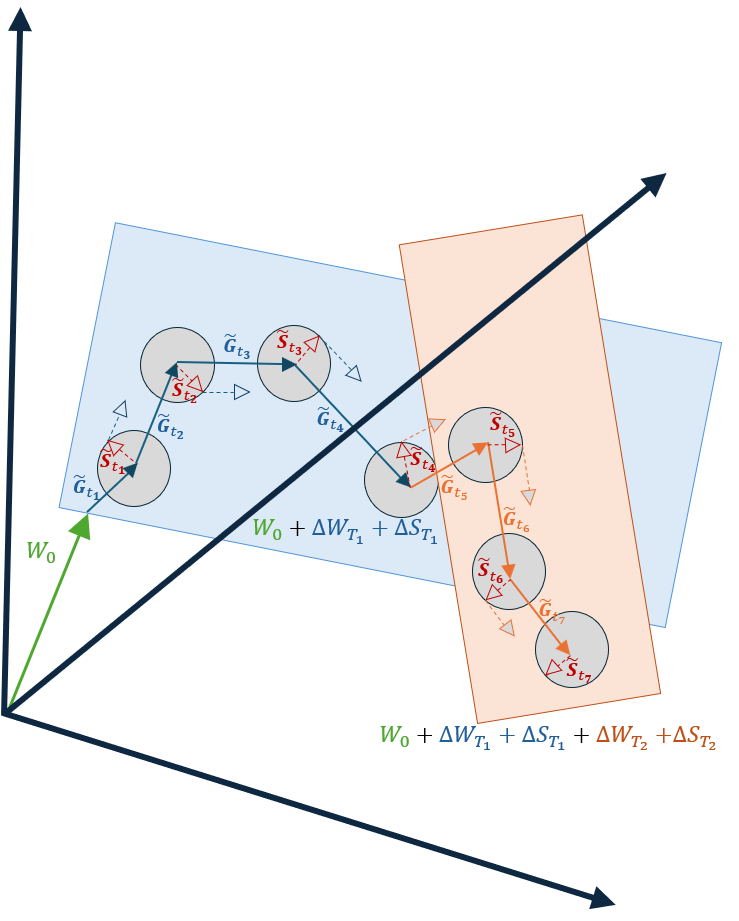} 
\caption{The illustration depicts the training process of LORENZA (\ref{alg:LORENZA}). The process begins by selecting a low-rank subspace using the efficient SSRF algorithm (\ref{alg::randomized_range_finder}), visualized here as a 2D plane (blue and orange). Next, a low-rank AdaZo-SAM optimization step (\ref{alg:AdaZo_SAM}) is performed. Specifically, the estimated low-rank ascent direction \(\Tilde{\bbs}_t\), is computed using the RGE method, on the 2D-subspace. This low-rank ascent direction is being used to calculate the adversarial gradient \(\bbg_t\), at the perturbated weights, $\bbw_t+\rho \frac{\Tilde{\bbs}_t}{\left\|\Tilde{\bbs}_t\right\|_2},$ then projected onto the 2D-subspace, namely as \({\Hat{\bbg}_t}^{2 \times m} = \bbq_t^{2 \times n} {\Tilde{\bbg}_t}^{n \times m}\). Following this, a low-rank Adam optimization step is applied. After a predetermined number of LORENZA steps, the optimization subspace \(\bbq_t\) is updated, and the process is repeated.}
\label{fig:wrapfig}
\end{figure}

\section{Method}

In this section, we propose a computationally and memory-efficient adaptive-SAM optimization method that utilizes efficient zero-order gradient estimation to compute the ascent direction (perturbation) and analyze its convergence guarantees. Next, we leverage a computationally efficient subspace selection method, which offers lower complexity compared to SVD decomposition. This subspace selection is later used to determine the optimization subspace (gradient projection) for the final LORENZA optimization. Finally, we present LORENZA, which applies memory and computationally efficient low-rank SAM optimization updates.

\subsection{Single gradient SAM approach via zeroth-order ascent estimation}

In this subsection, we introduce Algorithm \ref{alg:AdaZo_SAM}, which we call AdaZo-SAM. This algorithm is designed to reduce computational effort, training time, and memory consumption associated with the adaptive SAM schema. Unlike traditional gradient-based optimizers, SAM and its variants require the calculation of two gradients, effectively performing backpropagation twice in a single update step. Inspired by the randomized gradient estimator (RGE) \cite{nesterov2017random,duchi2015optimal}, which relies on the finite difference of function values along a randomly chosen direction vector, we propose estimating the perturbation or ascent direction instead of calculating the exact SAM perturbation. This approach eliminates the need for additional costly gradient calculations, allowing backpropagation to be applied only once during each update step rather than twice.

Given a scalar-valued function $f(\bbw)$ where $\mathbf \bbw \in \mathbb R^{m\times n}$, the RGE (referred to as $\hat{\nabla}f(\bbw)$) is expressed using a central difference scheme, namely
\begin{footnotesize}
\begin{align}
    \hat{\nabla}f(\bbw) = \frac{1}{q}\sum_{i=1}^q \left [ \frac{f(\bbw + \mu \bbu_i) - f(\bbw - \mu \bbu_i)}{2\mu}  \bbu_i \right ]
    \label{eq: RGE}
\end{align}
\end{footnotesize}
 where $\mathbf \bbu_i\in\mathbb{R}^{m\times n}$ is a random direction vector typically drawn from the standard Gaussian distribution $\mathcal N (\mathbf 0, \mathbf I)$, $q$ is the number of function queries, and $\mu > 0$ is a small perturbation stepsize (also known as smoothing parameter).

\begin{algorithm}[htb]
\caption{Efficient single gradient step Adaptive SAM with zeroth-order ascent estimation (AdaZo-SAM)}
\label{alg:AdaZo_SAM}
\begin{algorithmic}
\STATE {\bfseries Inputs:} Initial parameters $\bbw_0, \bbm_{-1}={\bbv}_{-1}=0$, 
base learning rate $\gamma$, neighborhood size $\mu$, perturbation size $\rho,$ and momentum parameters $\beta_1, \beta_2$, small number $q\in\mathbb{N}$ ({\color{blue}default $q=1$}).
\STATE {\bfseries Output:} Optimized parameter $\bbw_{T+1}.$
\FOR{iteration $t \in\{0,1,2, \ldots, T-1\}$} 
\STATE Sample mini-batch $B=\left\{\xi_{1}, \xi_{2}, \ldots, \xi_{{|B|}}\right\}$
\STATE Compute ascent direction (perturbation): \\
\STATE \: (1) Random $\bbu_i\sim\mathcal{N}(\mathbf{0}, \mathbf{I}),i\in[q]$
\STATE \: (2) $\bbg_t^{\text{Pert}} =\frac{1}{q} \sum_{i=1}^q\left[\frac{-f\left(\bbw+\mu \bbu_i;\xi_t\right)+f\left(\bbw-\mu \bbu_i;\xi_t\right)}{2 \mu}\bbu_i\right]$ 
\STATE Compute SAM gradient:\\
$\quad\bbg_t^{\text{SAM}} = \nabla_\bbw f(\bbw_t + \rho\frac{\bbg_t^{\text{Pert}}}{\left\|\bbg_t^{\text{Pert}}\right\|_F};\xi_t)$
\STATE Update momentum and variance:
\STATE $\quad\bbm_t  = \beta_1\bbm_{t-1} + \left(1 - \beta_1\right)\bbg_t^{\text{SAM}}$
\STATE $\quad\bbv_t  = \beta_2\bbv_{t-1} + \left(1 - \beta_2\right)(\bbg_t^{\text{SAM}})^2$
\STATE $\hat{\bbm}_t  = \bbm_t / \left(1 - \beta_1^t\right)$
\STATE $\hat{\bbv}_t = \bbv_t / \left(1 - \beta_2^t\right)$
\STATE Update parameters:
\STATE  $\bbw_{t+1}  = \bbw_t - \gamma \hat{\bbm}_t / \left(\sqrt{\hat{\bbv}_t} + \epsilon\right)$
\ENDFOR
\end{algorithmic}
\end{algorithm}   
It might be assumed that estimating the ascent direction would break the convergence guarantees of SAM and AdamSAM. However, following, we show that even using just one randomized matrix, i.e., \( q = 1 \), is sufficient to maintain the same convergence properties.
To analyze the convergence of the AdaZo-SAM algorithm, 
for the simplicity of writing, we let $ \mathbb{E}_\xi[\nabla f(\bbw)]=\mathbb{E}_\xi[\nabla_{\bbw} f(\bbw ; \xi)]$, where $\xi\sim \mathbb{P}_\mathcal{D}$ is a stochastic input batch, and $\mathbb{P}_\mathcal{D}$ is the sampling distribution over dataset/domain $\mathcal{D}.$

\begin{theorem}[AdaZo-SAM convergence rate]\label{theorem::AdaZo_SAM_convergence}
Consider a $\beta$-smooth, non-convex function $f$ parametrized by a matrix $\mathbf{W} \in \mathbb{R}^{m \times n},$ where $m \leq n,$ without loss of generality.  Suppose $f$ satisfying $\sup _{\bbw} \mathbb{E}_\xi\|f(\mathbf{W};\xi)\|\leq C$ for some large $C\in\mathbb{R}_+$ then, Algorithm \ref{alg:AdaZo_SAM} initialized at $\mathbf{W}_0$ with step size 
$\eta=\frac{1}{\beta\sqrt{T}},$ 
$$\frac{1}{T} \sum_{t=0}^{T-1} \mathbb{E}\left\|\hat{\nabla} f\left(\bbw_t\right)\right\|_F^2\leq \mathcal{O} \left(\frac{C\beta}{\sqrt{T}}\right) + \beta^2 \rho^2.$$
where $\hat{\nabla} f\left(\bbw_t\right)$ is the RGE (\ref{eq: RGE}) of function $f$ with $q=1,\mu\rightarrow0,$ and $\xi\sim \mathbb{P}_\mathcal{D}$ is a stochastic  batch, drawn by distribution $\mathbb{P}_\mathcal{D}$ over domain $\mathcal{D}.$  
\end{theorem}
The proof of Theorem \ref{theorem::AdaZo_SAM_convergence} can be found in Appendix~\ref{sec:A proof}. Similarly to all other variants of SAM (such as Theorem 3.5 in \cite{si2023} and Theorem 1 in \cite{sun2023adasamboostingsharpnessawareminimization}),  Theorem~\ref{theorem::AdaZo_SAM_convergence} shows that SAM with a decaying or sufficiently small perturbation size \(\rho\) converges to stationary points for non-convex smooth functions.
Practically, for example, GSAM suggest to schedule $\rho_t$ by $$\rho_t=\rho_{\min }+\frac{\left(\rho_{\max }-\rho_{\min }\right)\left(l r-l r_{\min }\right)}{l r_{\max }-l r_{\min }},$$
allowing to reach a flat stationery point area, then decaying the perturbation proportionally to the learning rate. 

\subsection{Efficient subspace selection} 
For a matrix \(\mathbf{A} \in \mathbb{R}^{n \times m}\), finding an effective low-rank approximation can be framed as solving the optimization problem $\min _{\mathbf{Q}, \mathbf{U}}\|\mathbf{A} - \mathbf{Q} \mathbf{U}^\top\|_F^2,$
where \(\mathbf{Q} \in \mathbb{R}^{n \times r}\) and \(\mathbf{U} \in \mathbb{R}^{m \times r}\). The resulting approximation, \(\mathbf{A}_{\text{app},r} = \mathbf{Q} \mathbf{U}^\top\), provides a rank-\(r\) representation of \(\mathbf{A}\). However, directly computing this via the singular value decomposition (SVD) becomes computationally prohibitive for large matrices due to its high complexity and memory requirements.  

As suggested in \cite{refael2025adarankgrad}, we utilize an efficient alternative inspired by the randomized range finder approach outlined in \citep{halko2010findingstructurerandomnessprobabilistic}. This method focuses on solving the problem $\arg\min _{\mathbf{Q} \in \mathbb{R}^{n \times r}} \|\mathbf{A} - \mathbf{Q} \mathbf{Q}^\top \mathbf{A}\|_F,$ where the low-rank approximation is constructed as \(\mathbf{A}_{\text{app},r} \approx \mathbf{Q} \mathbf{Q}^\top \mathbf{A}\).  

The computational efficiency of this method stems from its key operations: multiplying the matrix \(\mathbf{A}\) with a random projection matrix \(\boldsymbol{\Omega}\) (complexity \(O(mnr)\)) and performing a QR decomposition on the resulting matrix \(\mathbf{Y}\) (complexity \(O(mr^2)\)). The overall complexity, \(O(mnr + mr^2)\), simplifies to \(O(mnr)\) when \(r \ll m, n\), making it significantly more scalable than SVD, which requires \(O(\min(mn^2, m^2n))\). This approach enables efficient extraction of leading singular vectors in large-scale data, making it well-suited for computationally constrained scenarios.
 \begin{algorithm}[htb]
        \caption{Subspace selection via randomized range finder (SSRF)}\label{alg::randomized_range_finder}
        \begin{algorithmic}
            \STATE {\bfseries Inputs:} Matrix $\bba \in \mathbb{R}^{m \times n}$, with $m \leq n$, target rank $r\leq \min \{n, m\}$. 
            \STATE {\bfseries Initialization:} $\boldsymbol{\Omega} \in \mathbb{R}^{n\times r} \sim \mathcal{N}(0,1/r)$
            \STATE Compute $\bby \leftarrow \bba \boldsymbol{\Omega}$
            \STATE Construct $\bbq\in \mathbb{R}^{m\times r},\bbr\in\mathbb{R}^{r\times n}$ using the $\mathrm{QR}$ decomposition of $\bby$
            \STATE {\bfseries Return:} $\bbq,\bbr$ 
        \end{algorithmic}
    \end{algorithm}
    
\subsection{Low-rank gradient optimization
via efficient zeroth-order adaptive SAM}

\definecolor{highlightcolor}{rgb}{0.9, 0.9, 0.9}  %
\begin{algorithm}[htb]
   \caption{LORENZA: Low-rank gradient optimization via efficient zeroth-order AdaSAM}
   \label{alg:LORENZA}
 \begin{algorithmic}
   \STATE {\bfseries Input:} A weight matrix $\textbf{W} \in \mathbb{R}^{m \times n}$ with $m \leq n$. Step size $\eta$, scale factor $\alpha$, decay rates \{$\beta_1, \beta_2$\}, weight decay $\lambda$, rank $r$, subspace update frequency $T$, small number $q\in\mathbb{N}$ ({\color{blue}default $q=1$}), a small interval length $\mu$.
   \STATE \textbf{Initialize}: $t \gets 0$ and optionally, $\rho_t$ schedule: $\rho_t=\rho_{\min }+\frac{\left(\rho_{\max }-\rho_{\min }\right)\left(l r-l r_{\min }\right)}{l r_{\max }-l r_{\min }}$
   \REPEAT
   \STATE \textcolor{blue}{\# Block 1: Calculate low rank gradient projection.}
    \STATE Sample mini-batch $B=\left\{\xi_1, \xi_2, \ldots, \xi_{{|B|}}\right\}$
   \IF{$t \bmod T = 0$
   } 
    \STATE Compute $\bbg_t \leftarrow \frac{1}{|B|}\sum_{\xi_i \in B}\frac{\partial}{\partial \bbw} f( \bbw_t; \xi_i)$
    \STATE $\mathbf{Q}_t,\bbr_t\leftarrow\operatorname{SSRF}(\mathbf{G}_t, r)$ 
   \ENDIF \; {\color{gray}\# Alternatively criteria $\|\hat{\bbg}_{t}\|\leq\varsigma$}
   \STATE \textcolor{blue}{\# Block 2: Low-rank rank ascent perturbation}
    \STATE \textcolor{gray}{Compute low-rank random directions}
    \STATE 
    Randomize vector ${\bu_j} ^{r\times 1}\sim\mathcal{N}(\mathbf{0}, 1), \,\, j\in[q]$
    \STATE Set $\bbp_j= \mathbf{Q}_t\diag(\bu_j)\bbr_t, \,\, j\in[q]$
   \STATE {\color{gray}{Compute ascent direction (perturbation)}} \\
   \begin{footnotesize}
   $\bbg_t^{\text{Pert}} =-\frac{1}{q} \sum_{\xi_i \in B,j\in[q]}\left[\frac{f\left(\bbw_t+\mu\bbp_j; \xi_i\right)-f\left(\bbw_t-\mu \bbp_j; \xi_i\right)}{2 \mu} \bbp_j\right]$ 
   \end{footnotesize}
    \STATE \textcolor{blue}{\# Block 3: Low-rank adaptive SAM}
    \STATE {\color{gray}{Compute SAM gradient}}\\
    $\bbg^{\text{SAM}}_t = \frac{1}{|B|} \sum_{\xi_i \in B}\frac{\partial}{\partial \bbw} f\left( \bbw_t + \rho\frac{\bbg_t^{\text{Pert}}}{\|\bbg_t^{\text{Pert}}\|_F} ; \xi_i \right)$
    \STATE $\hat{\bbg}_t \longleftarrow \bbq_t^\top\bbg^{\text{SAM}}_t$
    \STATE $\bbm_t  \longleftarrow \beta_1 \bbm_t + \left(1 - \beta_1\right)\hat{\bbg}_t$
    \STATE $ \bbv_t  \longleftarrow \beta_2 \bbv_t + \left(1 - \beta_2\right) \hat{\bbg}_t^2$
    \STATE $\hat{\bbm}_t  \longleftarrow \bbm_t / \left(1 - \beta_1^t\right)$
    \STATE $\hat{\bbv}_t \longleftarrow \bbv_t / \left(1 - \beta_2^t\right)$
   \STATE \textcolor{blue}{\# Block 4: Update weight in original space.}
    \STATE $\bbw_t  \longleftarrow \bbw_t - \alpha \bbq_t\hat{\bbm}_t / \left(\sqrt{\hat{\bbv}_t} + \epsilon\right)$
   \STATE $t \gets t + 1$
   \UNTIL{convergence criteria met {\color{gray} (e.g. epoch number, gradient norm )}}
    \STATE \textbf{return} $\mathbf{W}_T$ \textcolor{gray}{\# A flat local minima}
 \end{algorithmic}
\end{algorithm}
In this subsection, we present our main algorithm, LORENZA, an efficient sharpness-aware and
adaptive low-rank gradients and moments optimization method.  
\begin{definition}\label{def:lor}
[Low-Rank Gradient by Efficient Zeroth-Order Adaptive SAM (LORENZA)]
LORENZA defines the following gradient update rules.
\begin{footnotesize}
\begin{equation*}
\textbf{LORENZA}\left\{
\begin{aligned}
&\bbg_t^{\text{Pert}} = \bbq_t \bbq_t^\top\Hat{\nabla}_\bbw f\left(\bbw_t ;\xi_t\right)\bbr_t\bbr_t^\top\quad \\
&\bbg_t^{\text{SAM}} = \nabla_\bbw f\left(\bbw_t + \rho\frac{\bbg_t^{\text{Pert}}}{\left\|\bbg_t^{\text{Pert}}\right\|_F};\xi_t\right)\\
&\hat{\bbg}_t=\bbq_t \rho_t\left(\bbq_t^\top \bbg_t^{\text{SAM}} \bbr_t\right) \bbr_t^\top\\
&\bbw_T=\bbw_0+\eta \sum_{t=0}^{T-1} \hat{\bbg}_t,
\end{aligned} 
\right.
\end{equation*}
\end{footnotesize}
where $\rho_t$ is an entry-wise stateful gradient regularizer (e.g., Adam), $\bbq_t \in \mathbb{R}^{m\times r}$ and $\bbr_t \in \mathbb{R}^{r \times n}$ are projection matrices, $T\in\mathbb{N}$ is the subspace update time, $\eta$ is the learning rate, and $\xi_t$ is a stochastic batch.
\end{definition}
Note that Definition \ref{def:lor} describes the LORENZA step in a general form. However, the perturbation projection is actually implemented more efficiently (as in Algorithm \ref{alg:LORENZA}).
\begin{table*}[htb]
\centering
\caption{Evaluating LORENZA (including gradient alignment
term), comparing to state-of-the-art memory-efficient fine-tuning methods on GLUE benchmark using pre-trained RoBERTa-Base. We used NVIDIA A100 for the experiments.}
\begin{tabular}{lccccccccc}
\hline
\textbf{Model}  & \textbf{Memory} & \textbf{CoLA} & \textbf{STS-B} & \textbf{MRPC} & \textbf{RTE} & \textbf{SST2} & \textbf{MNLI} & \textbf{QNLI} & \textbf{QQP} \\
\hline
Full Fine-Tuning  & 747M & 62.24 & 90.92 & 91.30 & 79.42 & 94.57 & 87.18 & 92.33 & 92.28 \\ \hline
$\nu$ SAM & $>747$M & 57.38& 87.43 &90.46 &59.93&  92.68 &84.16 &90.11& 91.29\\ \hline
SAM-ON & $>747$M & 57.38& 87.43 &90.46 &59.93&  92.68 &84.16 &90.11& 91.29\\ \hline
LoRA (rank=4) & 257M & 61.38 & 90.57 & 91.07 & 78.70 & 92.89 & 86.82 & 92.18 & 91.29 \\
GaLore (rank=4) & 253M & 60.35 & 90.73  & 92.25 & 79.42 & 94.0  & 87.0 & 92.24 & 91.06 \\ 
LORENZA (rank=4) & 253M & \textbf{61.51} & \textbf{91.01}  & \textbf{92.57} & \textbf{81.26} & \textbf{94.63}  & \textbf{87.2} & \textbf{92.54} & \textbf{91.82} \\ 
\hline
LoRA (rank=8) & 264M & 61.83 & 90.80 & 91.90 & 79.06 & 93.46 & 86.94 & 92.25 & 91.22 \\
GaLore (rank=8) & 257M & 60.06 & 90.82 & 92.0 & 79.78 & 94.38 & \textbf{87.17} & 92.2   & 91.11  \\
LORENZA (rank=8) & 257M & \textbf{62.1} & \textbf{90.93} & \textbf{92.8} & \textbf{81.17} & \textbf{94.84} & 87.14 & \textbf{92.72} & \textbf{91.26}  \\
\hline
\end{tabular}
\label{tab:comparison}
\end{table*}

Recently, \citep{zhao2024galore,refael2025adarankgrad} analyzed the gradient structure of a broad family of nonlinear networks known as \emph{reversible networks}\footnote{These networks are formally defined in Appendix~\ref{Reversibility}.} \citep{tian2021}. This family includes various types of layers, such as linear layers (MLP and convolutional) and (leaky) ReLU nonlinearities. It was proved that the low-rank structure of their gradients naturally diminishes during training and fine-tuning, a phenomenon observed empirically across various layer types. This insight inspired the development of optimization methods that leverage low-rank update steps to reduce memory usage while enhancing accuracy.

Consider a neural network denoted as $\Phi(\cdot;\boldsymbol\theta)$, which consists of $L$ layers and is parameterized by {\footnotesize{$\boldsymbol{\theta} \triangleq \left[\bbw_1^{d_1 \times d_0}, \ldots, \bbw_{L-1}^{d_{L-1} \times d_{L-2}}, \bbw_{L}^{d_{L} \times d_0^{L-1}}\right]$}}. Here, $\bbw_{i}$ represents the weights tensor parameters associated with the $i$-th layer, for $i \in [L]$. 
By $f(\boldsymbol{\theta} ; \xi)$ we denote the loss $\mathcal{L}$ for a sample $\xi$ and prediction $\Phi(\xi;\boldsymbol\theta)$.
With a slight abuse of notation, we write $f(\bbw ; \xi)$ if the context refers to the weights of a certain layer.
The proposed algorithm, LORENZA, is stated in Algorithm~\ref{alg:LORENZA}.
It comprises four main blocks, all contained within an outer loop that terminates once we reach convergence. The role of each block is as follows.
\begin{itemize}[leftmargin=*]
    \item \textbf{Block 1}: We select the subspace (approximated) along the directions of the $r$ largest eigenvectors, using Algorithm \ref{alg::randomized_range_finder}, where $r$ is predefined. This subspace is being updated every predefined constant number of optimization steps, or alternatively, by a low-rank gradient convergence criterion (i.e., the gradient projected on the low-rank subspace converged, namely $\|\hat{\bbg}_t\|\leq\varsigma$).   
    \item \textbf{Block 2}: We calculate the low-rank-perturbation representing the adversarial (ascent) direction within a low-rank subspace. This is achieved by randomizing a linear combination of directions within the subspace, projecting it back onto the original space, and then using it to empirically estimate the projected gradient on the selected subspace (i.e., the gradient components that reside in the selected low-rank subspace).
    Note that throughout the paper, we use $q=1$. 
    \item \textbf{Block 3}: The SAM direction is calculated using the low-dimensional and memory-efficient projected gradients on the selected subspace. It then updates Adam's estimates for the first and second moments.
    \item \textbf{Block 4}: The low-rank Adam step is being projected back onto the full dimension, and the model parameters are getting updated until convergence criteria are met. Such criteria could be, for example, the number of epochs or gradient norm reaching below a predefined threshold, namely $\left\|\bbg_t^j\right\|_F^2\leq\varepsilon.$)
\end{itemize}

\begin{theorem}[Convergence of LORENZA]\label{thm:Convergence_LORENZA}
Consider a $\beta$-smooth nonconvex composition of $f\equiv\mathcal{L}\left(\Phi(\cdot)\right)$ that is bounded by some $M\in\mathbb{R}_+$. Let $\bbg_t^j$ denote the gradient matrix w.r.t. the $j$-th reversible layer $\bbw_t^j,$ at time $t\in\mathbb{N}$, for all $j\in[L]$ and $t\in\mathbb{N}$, and $T_\ell,\ell\in\mathbb{N}$ times are set by a convergence criterion (that is, $\|\hat{\bbg}_{\mathsf{T}_\ell}\|\leq\varsigma_\ell$). Consider any decay perturbation $\rho$ then, for any $\varepsilon> 0$, there exist $\mathsf{C}\in\mathbb{R}_+$ and $N$ such that for all $\mathsf{T}_N>\frac{\mathsf{C}}{\varepsilon^2}$, $\frac{1}{\mathsf{T}_N}\sum_{i=0}^{N-1}\sum_{t=\mathsf{T}_{i}}^{\mathsf{T}_{i+1}-1}\left\|{\bbg_t^j}^{\text{SAM}}\right\|_F^2 \leq \varepsilon
$. Principally, Algorithm \ref{alg:LORENZA}, with vanilla SGD weight update\footnote{We focus on SGD for the simplicity (as is standard practice in related literature, e.g., \citep{zhao2024galore}).}, 
achieves an $\varepsilon$-critical point,\footnote{Also known as $\varepsilon$-stationary, see, e.g., \citep{cossonLowRankGradientDescent2023}.} i.e., $\left\|\bbg_t^j\right\|_F^2\leq\varepsilon$, for some $t\in\mathbb{N}$, and any $j\in[L]$.
\end{theorem}

The proof of Theorem \ref{theorem::AdaZo_SAM_convergence} can be found in Appendix~\ref{sec:A proof}. Several important points should be noted. First, to reduce memory usage, Algorithm \ref{alg:LORENZA} updates weights on a per-layer basis during backpropagation, following recent approaches (e.g., \cite{lv2024adalomolo}). This differs from standard optimizers, which typically store full gradients in memory before updating all weights, leading to inefficiencies. Second, the Adam update step in Algorithm \ref{alg:LORENZA} can be replaced with any quantized Adam variant (e.g., \cite{NIPS2017_1c303b0e,chen2021quantized,seok2021quantized}), enabling fine-tuned quantized models or quantized adaptors. Finally, 4-bit projected gradient updates, as introduced in Q-GaLore \citep{zhang2024qgalore}, can be easily incorporated.

\section{Experiments}
\paragraph{Fine-tuning on the GLUE Benchmark.}  
We evaluate our approach on the GLUE benchmark \citep{wang2019superglue} by fine-tuning the pre-trained RoBERTa-base model \cite{liu2019roberta}. The results, compared against full fine-tuning, LoRA, and GaLore methods, are summarized in Table \ref{tab:comparison}. For evaluation metrics, we report overall accuracy (matched and mismatched) for MNLI, Matthew’s correlation for CoLA, Pearson correlation for STS-B, F1-score for MRPC, and accuracy for the remaining tasks. Our method demonstrates improved fine-tuning accuracy while maintaining comparable training memory on average. We employ $\rho$ scheduling (as proposed in the GSAM method \cite{zhuang2022surrogate}), with $\rho_{max}=0.01$, $\rho_{min}=1\text{e}-6$, and a cosine annealing learning rate scheduler.

\paragraph{Pre-training LLAMA on C4 Dataset.}
We repeated the comparison from \cite{zhao2024galore} [Table 2] to evaluate the performance of LORENZA, comparing the state-of-the-art method in terms of perplexity and memory efficiency. For this evaluation, we trained large LLaMA-based models on the C4 dataset, a curated and extensive version of the Common Crawl web corpus \citep{raffel2020exploring}. This dataset is widely used for pre-training language models and developing word representations. To better reflect real-world pre-training scenarios, we conducted training on a non-repeating, large-scale dataset and scaled model sizes up to 350 million parameters. The results of these experiments are shown in Table~\ref{ex::pretraining}.
\begin{table}[htb]
\caption{Comparison of low-rank state-of-the-art algorithms for pre-training LLaMA models of varying sizes on the C4 dataset. The results are reported in terms of validation perplexity. Experiments were conducted using NVIDIA H200 GPU.}
\centering
\begin{tabular}{l@{\hskip 4pt}c@{\hskip 4pt}c@{\hskip 4pt}c}
\hline 
\textbf{Method} & \textbf{60M} & \textbf{130M} & \textbf{350M}  \\
\hline 
Full-Rank & $34.06$ ($0.36$G) & $25.08$ ($0.76$G) & $18.80$ ($2.06$G) \\
\hline 
GaLore    & $34.88$ ($0.24$G) & $25.36$ ($0.52$G) & $18.95$ ($1.22$G) \\
Low-Rank  & $78.18$ ($0.26$G) & $45.51$ ($0.54$G) & $37.41$ ($1.08$G) \\
LoRA      & $34.99$ ($0.36$G) & $33.92$ ($0.80$G) & $25.58$ ($1.76$G) \\
ReLoRA    & $37.04$ ($0.36$G) & $29.37$ ($0.80$G) & $29.08$ ($1.76$G)\\
LORENZA    & $\textbf{34.29}$ ($0.24$G) & $\textbf{24.92}$ ($0.53$G) & $\textbf{18.87}$ ($1.24$G) \\
\hline
\makecell{Training \\ Tokens}  & $1.1$B & $2.2$B & $6.4$B \\
$r / d_{\text{model}}$   & $128 / 256$ & $256 / 768$ & $256 / 1024$ \\
\hline
\end{tabular}
\label{ex::pretraining}
\end{table}

\begin{table}[htb]
\centering
\caption{The LLaMA 7B model was pre-trained on the C4 dataset for 120K steps with \underline{\textbf{8-bit quantization}} range in optimization. Adam was used for full-parameter training, while LORENZA and Galore were utilized for low-rank training, specifically with a rank of 256. Experiments were conducted using NVIDIA H200 GPU.}
\begin{tabular}{lccc}
\hline \textbf{Steps}/\textbf{Tokens} & \textbf{GaLore} & \textbf{Adam} & \textbf{LORENZA} \\
\hline 40\text{K }/ 5.2\text{B} & 17.94 & 18.09 & \textbf{17.81}\\
80\text{K }/10.5\text{B} & 15.39 & 15.47 & \textbf{15.28} \\
120\text{K }/15.7\text{B} & 14.95 & 14.83 & \bf{14.82} \\
\hline
\end{tabular}
\end{table}

\paragraph{Few/Zero-shot reasoning and long-context generalization.}
To evaluate our method performance on a complex reasoning task, we use the GSM8K dataset \cite{cobbe2021training}, testing systematic generalization. For these experiments, we used a batch size of 32 and 10 epochs for fine-tuning. We present the performance result in Table \ref{phi_gs8k} training Phi-2 (2.7B) model \cite{javaheripi2023phi}, and in Table \ref{lama_gs8k} training Lamma (1B) model \cite{touvron2023llama}. The results demonstrate that the proposed method significantly improves generalization to out-of-distribution data. The experiments were conducted on an NVIDIA H200 GPU.
\begin{table}[htb]
\centering
\caption{Zero shot evaluation performance on GSM8K dataset, training Phi-2 (2.7B).}
\begin{tabular}{lcc}
\hline
Phi-2 (2.7B)& Rank& Accuracy (0-shot) \\
\hline Base Model& 64& $15.16 \%$ \\
 Galore& 64& $52.24 \%$ \\
 LoRA& 64& $42.8 \%$ \\
 LORENZA& 64& $\textbf{53.37}$\%\\
 \hline\label{phi_gs8k}
\end{tabular}
\end{table}

\begin{table}[htb]
\centering
\caption{8-shot evaluation performance on GSM8K dataset, training  LLaMA (3B).}
\begin{tabular}{lcc}
\hline
LLaMA (1B) & Rank & Accuracy (8-shot) \\
\hline Base Model & 64& $17.93\%$ \\
 Galore & 64 & $74.9 \%$ \\
 LoRA & 64& $68.3 \%$ \\
 LORENZA & 64& $\textbf{76.4}$\%\\
 \hline\label{lama_gs8k}
\end{tabular}
\end{table}

\section{Discussion}
In this work, we introduce LORENZA, a novel memory-efficient optimization framework designed to enhance the generalization of parameter-efficient fine-tuning (PEFT) methods for large language models (LLMs). By combining zeroth-order sharpness-aware minimization (AdaZo-SAM) with low-rank gradient updates, LORENZA effectively reduces computational overhead while retaining the advantages of full-rank fine-tuning. Our approach demonstrates state-of-the-art performance across multiple benchmarks, outperforming existing low-rank adaptation methods such as LoRA and GaLore. 

Additionally, our theoretical analysis provides convergence guarantees, reinforcing the robustness of our method. Notably, LORENZA successfully bridges the gap between memory efficiency and generalization, making it an attractive alternative for resource-constrained training scenarios.

Our empirical results highlight LORENZA's superior accuracy, especially in challenging language tasks. Future work may explore adaptive rank selection, integration with quantization techniques, and evaluating effectiveness in knowledge editing \cite{rozner2024knowledge} or domain generalization \cite{roznerdomain}. By addressing key limitations in PEFT, LORENZA paves the way for more efficient and generalizable fine-tuning strategies, thereby advancing the broader field of LLM training and deployment.


\section*{Impact Statement}
This paper presents LORENZA, a novel optimization framework designed to enhance generalization in large language models (LLMs) through memory-efficient fine-tuning. By integrating zeroth-order sharpness-aware minimization with low-rank gradient updates, LORENZA significantly reduces the computational and memory overhead required for training and fine-tuning, making high-performance models more accessible on resource-constrained hardware. This contributes to the broader democratization of AI, enabling researchers and practitioners with limited computational resources to effectively fine-tune and deploy LLMs.

From an ethical perspective, our work aligns with the responsible advancement of machine learning by improving efficiency and accessibility. However, as with any method enhancing model adaptation, there is a need to consider the potential risks of misuse, including the fine-tuning of biased or harmful models. Future work should explore ways to integrate fairness-aware constraints and interpretability mechanisms into optimization frameworks like LORENZA to ensure ethical deployment. Overall, our contribution aims to push the boundaries of efficient deep learning while promoting responsible AI research and development.

\section*{Acknowledgments and Disclosure of Funding}
The work of TT is supported by the ISF grant No. 1940/23 and by MOST grant 0007091.

\bibliography{main.bib}
\bibliographystyle{icml2025}

\newpage
\appendix
\onecolumn
\section{Proofs}
\subsection{Proof Theorem \ref{theorem::AdaZo_SAM_convergence}}
\label{sec:A proof}
\emph{Consider a $\beta$-smooth, non-convex function $f$ parametrized by a matrix $\mathbf{W} \in \mathbb{R}^{m \times n},$ where $m \leq n,$ without loss of generality. Suppose $f$ satisfying $\sup _{\bbw} \mathbb{E}_\xi\|f(\mathbf{W};\xi)\|\leq C$ for some large $C\in\mathbb{R}_+$ then, Algorithm \ref{alg:AdaZo_SAM} initialized at $\mathbf{W}_0$ with step size 
$\eta=\frac{1}{\beta\sqrt{T}},$ 
$$\frac{1}{T} \sum_{t=0}^{T-1} \mathbb{E}\left\|\hat{\nabla} f\left(\bbw_t\right)\right\|_F^2\leq \mathcal{O} \left(\frac{C\beta}{\sqrt{T}}\right) + \beta^2 \rho^2.$$
where $\hat{\nabla} f\left(\bbw_t\right)$ is the RGE (\ref{eq: RGE}) of function $f$ with $q=1,\mu\rightarrow0,$ and $\xi\sim \mathbb{P}_\mathcal{D}$ is a stochastic  batch, drawn by distribution $\mathbb{P}_\mathcal{D}$ over domain $\mathcal{D}.$   
} 
\begin{proof}

First, consider the following notation. For the simplicity of writing, we let $\nabla f(\bbw)=\nabla_{\bbw} f(\bbw ; \xi)$, where $\xi\sim \mathbb{P}_\mathcal{D}$ is a stochastic input batch, and $\mathbb{P}_\mathcal{D}$ is the sampling distribution over dataset/domain $\mathcal{D}.$ Accordingly, we denote $\mathbb{E}\left[\nabla f(\bbw)\right]=\mathbb{E}_{\xi}\left[\nabla_{\bbw} f(\bbw ; \xi)\right]$. We denote the estimated gradient at $\bbw$ by
\begin{align}\mathbb{E}_\xi\left[\hat{\nabla}_\bbw f(\bbw;\xi)\right] = \frac{1}{q}\sum_{i=1}^q \mathbb{E}_\xi\left[ \frac{f(\bbw + \mu \bbu_i;\xi) - f(\bbw - \mu \bbu_i;\xi)}{2\mu}   \bbu_i \right ]\in\mathbb{R}^{m\times n},\nonumber
\end{align}
where $\bbu_i^{m\times n} \sim \mathcal{N}(\mathbf{0}, 1/n)$ is a randomized matrix. Similarly, for simplicity of writing, we denote $\hat{\nabla} f(\bbw)=\mathbb{E}_\xi\left[\hat{\nabla}_\bbw f(\bbw;\xi)\right].$ Notice that as $\mu \rightarrow 0$ and $q=1$, the finite difference of the function values in approaches $f^{\prime}(\bbw_t, \bbu_i):= \operatorname{Tr}\left(\nabla f(\bbw_t)^\top\bbu_i\right)$, denoting the directional derivative of $f(\bbw),$ along the random direction $\bbu_i,$ yielding $\hat{\nabla} f(\bbw_t) \rightarrow f^{\prime}(\bbw_t, \bbu_i) \bbu_i,$ thus,  
\begin{equation} \label{directional_derivative_rge}
\lim_{\mu \rightarrow 0}\mathbb{E}\left\|\hat{\nabla}f\left(\bbw_t\right)\right\|_F^2=\mathbb{E}_{\bbu}\left\| f^\prime\left(\bbw_t\right)\bbu_i\right\|_F^2 =\mathbb{E}_{\bbu}\left\|\operatorname{Tr}\left(\nabla f(\bbw_t)^\top\bbu_i\right)\bbu_i\right\|_F^2 = \mathbb{E}\left\|\nabla f\left(\bbw_t\right)\right\|_F^2.
 \end{equation}
For the simplicity of writing, let $\bbx_t=\bbw_t+\rho \frac{\hat{\nabla} f\left(\bbw_t\right)}{\left\|\hat{\nabla} f\left(\bbw_t\right)\right\|}$, thus $\bbw_{t+1}=\bbw_t-\eta \nabla f\left(\bbx_t\right)$. 
Now, by the $\beta$-smoothness of $f$, we have 
\begin{align} \mathbb{E} f\left(\bbw_{t+1}\right) \leq & \mathbb{E} f\left(\bbw_t\right)+ \mathbb{E}\left[\text{vec}\left(\nabla f\left(\bbw_t\right)\right)^\top\text{vec}\left(\bbw_{t+1}-\bbw_{t}\right)\right]+\frac{\beta}{2} \mathbb{E}\left\|\bbw_{t+1}-\bbw_{t}\right\|_F^2 \nonumber\\
\underset{(\rom{1})}{=} & \mathbb{E} f\left(\bbw_t\right)-\eta \mathbb{E}\left[\text{vec}\left(\nabla f\left(\bbw_t\right)\right)^\top\text{vec}\left(\nabla f\left(\bbx_t\right)\right)\right]+\frac{\beta \eta^2}{2} \mathbb{E}\left\|\nabla f\left(\bbx_t\right)\right\|_F^2 \nonumber\\
= & \mathbb{E} f\left(\bbw_t\right)-\frac{\eta}{2} \mathbb{E}\left\|\nabla f\left(\bbw_t\right)\right\|_F^2-\frac{\eta}{2} \mathbb{E}\left\|\nabla f\left(\bbx_t\right)\right\|_F^2+\frac{\eta}{2} \mathbb{E}\left\|\nabla f\left(\bbw_t\right)-\nabla f\left(\bbx_t\right)\right\|_F^2
\nonumber\\ & +\frac{\beta \eta^2}{2}\mathbb{E}\left\|\nabla f\left(\bbx_t\right)\right\|_F^2 \nonumber\\
\underset{(\rom{2})}{\leq} & \mathbb{E} f\left(\bbw_t\right)-\frac{\eta}{2} \mathbb{E}\left\|\nabla f\left(\bbw_t\right)\right\|_F^2+\frac{\beta^2 \eta}{2} \mathbb{E}\left\|\bbw_t-\bbx_t\right\|_F^2 \nonumber\\
\underset{(\rom{3})}{=} & \mathbb{E} f\left(\bbw_t\right)-\frac{\eta}{2} \mathbb{E}\left\|\nabla f\left(\bbw_t\right)\right\|_F^2+\frac{\beta^2 \rho^2 \eta}{2}\label{in_proof}\\ 
\underset{(\rom{4})}{=} & \mathbb{E} f\left(\bbw_t\right)-\frac{\eta}{2} \mathbb{E}\left\| \hat{\nabla} f\left(\bbw_t\right)\right\|_F^2+\frac{\beta^2 \rho^2 \eta}{2}\nonumber,\end{align}
where $(\rom{1})$ follows by the definition of $\bbx_t,$ $(\rom{2})$ follows from $\frac{\eta}{2}=\frac{1}{2 \beta \sqrt{T}} \geq \frac{1}{2 \beta T} = \frac{\beta\eta^2}{2}$, $(\rom{3})$
follows by $\left\|\bbx_t-\bbw_t\right\|_F=\left\|\rho \frac{\hat{\nabla} f\left(\bbw_t\right)}{\left\|\hat{\nabla} f\left(\bbw_t\right)\right\|}\right\|_F=\rho$, and finally $(\rom{4})$ follows by Equation (\ref{directional_derivative_rge}). Rearrearage both sides to bound the gradient Forbinus norm, we obtain
$$
\mathbb{E}\left\|\hat{\nabla} f\left(\bbw_t\right)\right\|_F^2 \leq \frac{2}{\eta}\left(\mathbb{E} f\left(\bbw_{t
}\right)-\mathbb{E} f\left(\bbw_{t+1}\right)\right)+\beta^2 \rho^2.
$$
Adding up the inequality for $t\in [T-1]$, and dividing both sides by $T$, we get
$$
\begin{aligned}
\frac{1}{T} \sum_{t=0}^{T-1} \mathbb{E}\left\|\hat{\nabla} f\left(\bbw_t\right)\right\|_F^2 & \leq \frac{2}{\eta T}\left(\mathbb{E} f\left(\bbw_0\right)-\mathbb{E} f\left(\bbw_T\right)\right)+\beta^2 \rho^2 \\
& \leq \frac{2 C}{\eta T}+\beta^2 \rho^2
\end{aligned}
$$

Now, choosing $\eta=\frac{1}{\beta\sqrt{T}}$, 
we have
$$  \frac{2 C}{\eta T}+\beta^2 \rho^2 \leq \frac{2C\beta}{\sqrt{T}}+\beta^2 \rho^2,$$
thus
$$\frac{1}{T} \sum_{t=0}^{T-1} \mathbb{E}\left\|\hat{\nabla} f\left(\bbw_t\right)\right\|_F^2\leq \mathcal{O} \left(\frac{C\beta}{\sqrt{T}}\right) + \beta^2 \rho^2.$$

\end{proof}

\subsection{Proof of Theorem~\ref{thm:Convergence_LORENZA}}\label{sec:B proof}
\emph{Consider a $\beta$-smooth nonconvex composition of $f\equiv\mathcal{L}\left(\Phi(\cdot)\right)$ that is bounded by some $M\in\mathbb{R}_+$. Let $\bbg_t^j$ denote the gradient matrix w.r.t. the $j$-th reversible layer $\bbw_t^j,$ at time $t\in\mathbb{N}$, for all $j\in[L]$ and $t\in\mathbb{N}$, and $T_\ell,\ell\in\mathbb{N}$ times are set by a convergence criterion (that is, $\|\hat{\bbg}_{\mathsf{T}_\ell}\|\leq\varsigma_\ell$). Consider any decay perturbation $\rho$ then, for any $\varepsilon> 0$, there exist $\mathsf{C}\in\mathbb{R}_+$ and $N$ such that for all $\mathsf{T}_N>\frac{\mathsf{C}}{\varepsilon^2}$, $\frac{1}{\mathsf{T}_N}\sum_{i=0}^{N-1}\sum_{t=\mathsf{T}_{i}}^{\mathsf{T}_{i+1}-1}\left\|{\bbg_t^j}^{\text{SAM}}\right\|_F^2 \leq \varepsilon
$. Principally, Algorithm \ref{alg:LORENZA}, with vanilla SGD weight update\footnote{We focus on SGD for the simplicity (as is standard practice in related literature, e.g., \citep{zhao2024galore}).}, 
achieves an $\varepsilon$-critical point,\footnote{Also known as $\varepsilon$-stationary, see, e.g., \citep{cossonLowRankGradientDescent2023}.} i.e., $\left\|\bbg_t^j\right\|_F^2\leq\varepsilon$, for some $t\in\mathbb{N}$, and any $j\in[L]$.}
\begin{proof} We denote by $\mathsf{T}_\ell\in\mathbb{N}$ the time index $t$ at which we update the subspace, at Block $1$ of the algorithm, for the $\ell$-th time, for $\ell\in\mathbb{N}$. For the simplicity of writing, for the $j$-th layer $\bbw_j,$ we omit $j$ from ${\bbw^j}$, and use instead $\bbg_t^j=\nabla_{\boldsymbol{\bbw^j} }f\left(\boldsymbol{\theta}_t\right)=\nabla f\left(\bbw_t\right)$. In addition, for simplicity, we let $\bbx_t=\bbw_t+\rho \frac{\nabla f\left(\bbw_t\right)}{\left\|\nabla f\left(\bbw_t\right)\right\|}$, thus $\bbw_{t+1}=\bbw_t-\eta \nabla f\left(\bbx_t\right)$. Consider the SVD decomposition of the gradient $\nabla_{\boldsymbol{\bbw^j} }f\left(\boldsymbol{\theta}_{\mathsf{T}_i}\right)= \bbu_{\mathsf{T}_i}\Sigma_{\mathsf{T}_i}\bbv_{\mathsf{T}_i}^\top$. Accordingly, for $t\in[\mathsf{T}_i,\mathsf{T}_{i+1}-1]$, we define the low rank gradient as $\hat{\bbg}_t \triangleq \bbq_{\mathsf{T}_i}\bbg_t,$ for $\bbq_{\mathsf{T}_i}= \bbu_{\mathsf{T}_i}\left[:,:r\right]\bbu_{\mathsf{T}_i}\left[:,:r\right]^\top,$ where $\bbu_{\mathsf{T}_i}$ is obtained by the subspace search, presented in the SSRF Algorithm~\ref{alg::randomized_range_finder}, using the exact truncated SDV calculation. Now, let $h_t\triangleq \mathbb{E} f\left(\bbw_t\right)-\mathbb{E} f\left(\bbw_{\mathsf{T}_i+1}\right)$, and $\eta\equiv\eta_t$ denote the learning rate. Then,
\begin{align} 
h_{t+1} &=   \mathbb{E} f\left(\bbw_{t+1}\right)-\mathbb{E} f\left(\bbw_{\mathsf{T}_i+1}\right)\nonumber \\
&\underset{(\rom{1})}{\leq} \mathbb{E} f\left(\bbw_t\right)-\mathbb{E} f\left(\bbw_{\mathsf{T}_i+1}\right)-\frac{\eta}{2} \mathbb{E}\left\|\nabla f\left(\bbw_t\right)\right\|_F^2+\frac{\beta^2 \rho^2 \eta}{2}
\nonumber \\
&\underset{(\rom{2})}{=} h_t-\frac{\eta}{2} \mathbb{E}\left\| \hat{\nabla} f\left(\bbw_t\right)\right\|_F^2+\frac{\beta^2 \rho^2 \eta}{2},
\label{eqn:upp}
\end{align}

where $(\rom{1}),$ follows Equation (\ref{in_proof}), and $(\rom{2})$ follows by Equation (\ref{directional_derivative_rge}).
Rearranging \eqref{eqn:upp}, and choosing $\eta_t=\eta$, for all $t\geq0$, we readily obtain that,
\begin{align}
\sum_{t=\mathsf{T}_{i}}^{\mathsf{T}_{i+1}-1}\mathbb{E}\left\|\hat{\nabla} f\left(\bbw_t\right)\right\|_F^2&\leq \frac{2(h_{\mathsf{T}_i}-h_{\mathsf{T}_{i+1}})}{\eta}+(\mathsf{T}_{i+1}-\mathsf{T}_i)\beta^2 \rho^2.\label{eqn:arg1} \nonumber 
 \end{align}
 Thus, for $N\in\mathbb{N}$,
\begin{align}
\frac{1}{\mathsf{T}_N}\sum_{i=0}^{N-1}\sum_{t=\mathsf{T}_{i}}^{\mathsf{T}_{i+1}-1}\mathbb{E}\left\|\hat{\nabla} f\left(\bbw_t\right)\right\|_F^2 &\leq \frac{1}{\mathsf{T}_N}\sum_{i=0}^{N-1}\left[\frac{2(h_{\mathsf{T}_i}-h_{\mathsf{T}_{i+1}})}{\eta}+(\mathsf{T}_{i+1}-\mathsf{T}_i)\beta^2 \rho^2 \right] \nonumber\\ 
& = \frac{2(h_{\mathsf{T}_0}-h_{\mathsf{T}_N})}{\eta \mathsf{T}_N}+\frac{(\mathsf{T}_N-\mathsf{T}_0)\beta^2 \rho^2 }{\mathsf{T}_N} \nonumber\\ 
&\leq\frac{M}{\beta\sqrt{\mathsf{T}_N}} +\beta^2 \rho^2,
\end{align}

where $\eta = \frac{1}{\beta\sqrt{\mathsf{T}_N }}$. Now by the definition of $\bbq_{\mathsf{T}_i}$ for any $i\in\mathbb{N}$ there exists some $\alpha\in (0,1]$, for which
\begin{equation}\label{info_thresh}
    \left\|\hat{\nabla} f\left(\bbw_{\mathsf{T}_i}\right)-\bbq_{\mathsf{T}_i}\hat{\nabla} f\left(\bbw_{\mathsf{T}_i}\right)\right\|_F^2\leq\alpha\left\|\hat{\nabla} f\left(\bbw_{\mathsf{T}_i}\right)\right\|_F^2. \end{equation}
Obviously, the following clearly holds for any $t\in\mathbb{N}$,  
\begin{align}
\left\|\hat{\nabla} f\left(\bbw_t\right)\right\|_F^2 &=\left\|\bbq_{\mathsf{T}_i}\hat{\nabla} f\left(\bbw_t\right)\right\|_F^2 + \left\|\bbq^\perp_{\mathsf{T}_i}\hat{\nabla} f\left(\bbw_t\right)\right\|_F^2\nonumber\\
&=\left\|\bbq_{\mathsf{T}_i}\hat{\nabla} f\left(\bbw_t\right)\right\|_F^2 + \left\|\hat{\nabla} f\left(\bbw_t\right)-\bbq_{\mathsf{T}_i}\hat{\nabla} f\left(\bbw_t\right)\right\|_F^2,\label{identity_proj}
\end{align}
and thus by plugging \eqref{info_thresh} into \eqref{identity_proj}, at $t=\mathsf{T}_i$, for any $i\in\mathbb{N}$, we get,
$(1-\alpha)\|\hat{\nabla} f\left(\bbw_{\mathsf{T}_i}\right)\|_F^2\leq\left\|\bbq_{\mathsf{T}_i}\hat{\nabla} f\left(\bbw_{\mathsf{T}_i}\right)\right\|_F^2.$
 Accordingly, 
\begin{align}\left\|\bbq^\perp_{\mathsf{T}_i}\hat{\nabla} f\left(\bbw_{\mathsf{T}_i}\right)\right\|_F^2\leq\frac{\alpha}{1-\alpha}\left\|\bbq_{\mathsf{T}_i}\hat{\nabla} f\left(\bbw_{\mathsf{T}_i}\right)\right\|_F^2.\label{7}
\end{align}

Recall from Lemma B.$3$, Equation $31,$ in \cite{zhao2024galore} that for the reversible layer, 
\begin{align}
\mathbb{E}\left\|\hat{\nabla} f\left(\bbw_t\right)\right\|_F^2 & =\mathbb{E}\left\|(I-\eta \bbs) \hat{\nabla} f\left(\bbw_{t-1}\right)\right\|_F^2\\
&\leq\left\|(I-\eta \bbs)\right\|_2^2 \mathbb{E}\left\|\hat{\nabla} f\left(\bbw_{t-1}\right)\right\|_F^2\\
&=\max_i|1-\eta\lambda_{i}|^2\mathbb{E}\left\|\hat{\nabla} f\left(\bbw_{t-1}\right)\right\|_F^2,
\end{align}
where $\{\lambda_i\}_i$ are the eigenvalue of $\bbs$. Thus, using the fact that $\bbs$ is positive semi-definite matrix,
If the learning rate $\eta$ satisfies $\eta \leq \frac{2}{\lambda_{\max}}$, where $\lambda_{\max}$ is the maximum eigenvalue of $\bbs$, it follows that $\max_i |1 - \eta\lambda_i|^2 \leq 1$. Consequently, this implies that $\mathbb{E}\left\|\hat{\nabla} f\left(\bbw_t\right)\right\|_F^2 \leq \mathbb{E}\left\|\hat{\nabla} f\left(\bbw_{t-1}\right)\right\|_F^2$. Therefore, the Frobenius norm of the gradient decreases monotonically as a function of time $t$.
Now, recall that $\varsigma_{i}$ is any positive number such that $\varsigma_{i}<\sqrt{1-\alpha}\cdot\|\hat{\nabla} f\left(\bbw_{\mathsf{T}_{i-1}}\right)\|_F$. According to (\ref{7}), this necessarily implies that in each block $i$, we will execute (at least once) the low-rank optimization block (indeed, the condition $\|\hat{\nabla} f\left(\bbw_t\right)_{\mathsf{T}_i}\|_F>\varsigma_{i}$ is satisfied). This, conjugated with the monotonicity property that $\left\|\hat{\nabla} f\left(\bbw_{t}\right)\right\|_F^2\leq\left\|\hat{\nabla} f\left(\bbw_{{\mathsf{T}_{i}}}\right)\right\|_F^2$, for any $t\in[\mathsf{T}_i,\mathsf{T}_{i+1}-1]$ and $i\in[N]$, imply that
\begin{align}
   \frac{1}{\mathsf{T}_N}\sum_{i=0}^{N-1}\sum_{t=\mathsf{T}_{i}}^{\mathsf{T}_{i+1}-1}\left\|\hat{\nabla} f\left(\bbw_t\right)\right\|_F^2&\leq \frac{1}{\mathsf{T}_N}\sum_{i=0}^{N-1}\sum_{t=\mathsf{T}_{i}}^{\mathsf{T}_{i+1}-1}\left\|\hat{\nabla} f\left(\bbw_{{\mathsf{T}_{i}}}\right)\right\|_F^2\\
   &\leq\frac{1}{(1-\alpha)\mathsf{T}_N}\sum_{i=1}^{N-1}\sum_{t=\mathsf{T}_{i}}^{\mathsf{T}_{i+1}-1}\left\|\bbq_{\mathsf{T}_{i}}\hat{\nabla} f\left(\bbw_t\right)\right\|_F^2\\
   &\leq \frac{M}{(1-\alpha)\beta\sqrt{\mathsf{T}_N}} +\frac{\beta^2 \rho^2}{1-\alpha}. 
\end{align}
Accordingly, for decaying perturbation $\rho\equiv\rho_t,$ without the loss of generality for any small enough $\varepsilon_1,\varepsilon_2\geq 0,$ there exist $\varepsilon\geq\varepsilon_1+\varepsilon_2$, where $\mathsf{T}_N>\frac{M^2}{(1-\alpha)^2\varepsilon_1^2}\geq\frac{\mathsf{C}}{\varepsilon^2}$, and $\rho_N<\varepsilon_2\frac{1-\alpha}{\beta^2},$  $$\min_{0\leq t \leq \mathsf{T}_{N}} {\left\|\hat{\nabla} f\left(\bbw_t\right)\right\|_F^2} \leq \frac{1}{\mathsf{T}_N}\sum_{i=0}^{N-1}\sum_{t=\mathsf{T}_{i}}^{\mathsf{T}_{i+1}-1}\left\|\hat{\nabla} f\left(\bbw_t\right)\right\|_F^2 \leq \varepsilon_1+\varepsilon_2\leq\varepsilon$$
and thus, there exists an iteration index $t\in[0,\mathsf{T}_{N}]$ for which,
\begin{align}
\left\|\hat{\nabla} f\left(\bbw_t\right)\right\|_F^2 \leq \varepsilon,
\end{align}
which, by definition, implies that Algorithm~\ref{alg:LORENZA} achieves an $\varepsilon$-critical point.

\end{proof}

\section{Additional definitions}
\begin{definition} (Reversibility \citep{tian2021}) \label{Reversibility} A neural network
$\phi$ that maps input $\boldsymbol{x}$ to output $\boldsymbol{y}=\phi(\boldsymbol{x};\theta)$ is reversible, if there exists $L(\boldsymbol{x} ;\theta)$ so that $\boldsymbol{y}=L(\boldsymbol{x} ;\theta) \boldsymbol{x}$, and the backpropagated gradient $\boldsymbol{g}_{\boldsymbol{x}}$ satisfies $\boldsymbol{g}_{\boldsymbol{x}}=L^{\top}(\boldsymbol{x};\theta) \boldsymbol{g}_{\boldsymbol{y}}$, where $\boldsymbol{g}_{\boldsymbol{y}}$ is the backpropagated gradient at the output $\boldsymbol{y}$. $L(\boldsymbol{x} ;\theta) $ depends on the input $\boldsymbol{x}$ and weight $\theta$ in the network $\phi$.
\end{definition}

\end{document}